\RequirePackage{fix-cm}
\documentclass[twocolumn]{svjour3}
\pdfoutput=1
\smartqed 
\usepackage{graphicx}
\usepackage{array}
\usepackage{float}
\usepackage[export]{adjustbox}
\usepackage{indentfirst}
\usepackage{amsmath}
\usepackage{hyperref}

\usepackage{latexsym}
\usepackage{times}
\usepackage{epsfig}
\usepackage{epstopdf}
\usepackage{amssymb}
\usepackage{algorithm2e}
\usepackage{gensymb}
\usepackage{xcolor}
\usepackage{cite}
\usepackage{marvosym}
\usepackage{color}
\usepackage{amsmath,amssymb,amsfonts}
\usepackage{algorithmic}
\usepackage{graphicx}
\usepackage{textcomp}
\usepackage{xcolor}
\usepackage[utf8]{inputenc} 
\usepackage[T1]{fontenc}    
\usepackage{hyperref}       
\usepackage{url}            

\begin{document}

\title{MelNet: A Real-Time Deep Learning Algorithm for Object Detection
}

 

\author{Yashar Azadvatan  \textsuperscript{1}          \and
        Murat Kurt  \textsuperscript{2} 
}


\institute{Yashar Azadvatan \at
                \textsuperscript{1} International Computer Institute, Ege University, {I}zmir, Turkey\\
            \email{yashar.azadvatan@gmail.com}           
           \and
           Murat Kurt \at
              \textsuperscript{2} International Computer Institute, Ege University, {I}zmir, Turkey\\
              \email{murat.kurt@ege.edu.tr}  
}

\date{Received: date / Accepted: date}

\maketitle

\begin{abstract} In this study, a novel deep learning algorithm for object detection, named MelNet, was introduced. MelNet underwent training utilizing the KITTI dataset for object detection. Following 300 training epochs, MelNet attained an mAP (mean average precision) score of 0.732. Additionally, three alternative models —YOLOv5, EfficientDet, and Faster-RCNN-MobileNetv3— were trained on the KITTI dataset and juxtaposed with MelNet for object detection.

The outcomes underscore the efficacy of employing transfer learning in certain instances. Notably, preexisting models trained on prominent datasets (e.g., ImageNet, COCO, and Pascal VOC) yield superior results. Another finding underscores the viability of creating a new model tailored to a specific scenario and training it on a specific dataset. This investigation demonstrates that training MelNet exclusively on the KITTI dataset also surpasses EfficientDet after 150 epochs. Consequently, post-training, MelNet's performance closely aligns with that of other pre-trained models.

\end{abstract}
\newpage
\section{Introduction}
\label{intro}

Individuals observe an image and promptly identify the objects it contains, discern their positions, and comprehend their interactions. The human visual system exhibits rapidity and precision, allowing for intricate activities such as driving a vehicle with minimal conscious deliberation \cite{RedmonetalYOLO2016}.

In recent years, deep learning has shown a tremendous growth to solve the challenging problems such as object detection, semantic segmentation, person re-identification, image retrieval, anomaly detection, skin disease diagnosis, and many more. Various types of neural networks have been defined in deep learning to learn abstract features from data, such as Multilayer Perceptron (MLP), Convolutional Neural Networks (CNN), Recurrent Neural Networks (RNN), and Generative Adversarial Networks (GAN). The important aspects of neural networks include weight initialization, loss functions, different layers, overfitting, and optimization\cite{DubeyetalNeuro2022}.

Object detection, powered by deep learning, refers to the field of computer vision that aims to enable computers to see and perceive objects, similar to humans. It involves the task of accurately detecting and recognizing objects in visual data. By leveraging deep neural networks, these models can learn complex features and patterns from large amounts of labeled data, enabling them to effectively identify and classify objects in real-world scenes. Detection \cite{Fidler2013} \cite{Kleban2008} involves two primary procedures: classification and localization. In the classification aspect \cite{Heetal2015}, the task is to predict the vehicle class depicted in the image or video. Meanwhile, in localization \cite{Uijlings2013}, the goal is to ascertain the positions of identified vehicles and showcase them by outlining a bounding box. Technological progress, including the utilization of high-resolution cameras and high-speed hardware devices, has become integral to such systems. Optimizing green signals at traffic intersections based on traffic density can be achieved, contributing to the alleviation of traffic congestion and a reduction in overall vehicle waiting times.

Vision-centric algorithms for vehicle detection fall into three categories, which are distinguished by their reliance on motion, handcrafted features, and Convolutional Neural Network (CNN) approaches. Motion-based techniques encompass optical flow, background subtraction, and frame subtraction. These methods focus on classifying and detecting moving vehicles by analyzing images or videos captured from stationary cameras. Methods based on handcrafted features include Histogram of Oriented Gradients (HOG) \cite{Girshick2015}, Scale-Invariant Feature Transform (SIFT) \cite{Linetal2017}, and Harr-like \cite{Renetal2015}. These approaches exhibit a limited feature representation. Despite their effectiveness with small data sizes and their independence from special hardware requirements, these methods necessitate feature extraction by experts in machine learning. The algorithm's efficiency is heavily contingent on the precision of the methods used for identifying and extracting features. Deep Learning (DL) models demand a substantial volume of training data to learn high-level features, mitigating the need for intricate feature extraction algorithms. However, these algorithms heavily rely on powerful machines, given the numerous matrix multiplication operations they perform. Graphical Processing Units (GPUs) prove to be highly efficient in optimizing these operations \cite{Mittal2022}.
 
 In the realm of computer vision, the task of object detection holds a central role, encompassing the identification of distinct classes such as humans, animals, or cars within digital images. The core objective of object detection is to engineer computational models and methodologies that deliver a fundamental data element crucial for computer vision applications: the precise delineation and localization of objects. The pivotal metrics of prime significance within the domain of object detection encompass accuracy, covering both the precision of classification and localization, in addition to the aspect of speed \cite{ZouetalarXiv2023}. Initially, object detection was trained to identify objects in visual images, and later real-time object detection became of great importance and has played a crucial role in our advancements. When it comes to real-time object detection, the speed and accuracy of the model are of paramount importance.
 
Beyond the common challenges encountered in other computer vision endeavors, like variations in object viewpoints, lighting conditions, and intraclass diversity, object detection introduces its own set of challenges. These encompass a range of factors, including but not confined to: coping with object rotation and scale variations (notably with small objects), achieving precise object localization, detecting densely packed and partially obscured objects, optimizing detection speed, and more \cite{ZouetalarXiv2023}.
 
Swift and precise algorithms designed for object detection pave the way for remarkable applications. These applications encompass enabling computers to autonomously operate vehicles devoid of dedicated sensors, furnishing real-time scene insights to aid human users, and unlocking the capabilities of versatile and responsive robotic systems \cite{RedmonetalYOLO2016}.
 
The evolution of this field continues to gain momentum, with research in this domain expanding ceaselessly, propelled by human necessities. The sheer marvel lies in the fact that computers possess the capacity to visually interpret images and comprehend objects within them. This phenomenon is truly remarkable. Companies such as Tesla and Volvo are harnessing these breakthroughs to offer invaluable services to their clientele through autonomous vehicles, thereby actualizing some long-held dreams, albeit partially, for people around the world.
 
Vehicle detection serves the purpose of identifying and precisely classifying vehicles within a monitored area, thereby determining their specific positions \cite{LiuetalSSD2016}. Object detection remains a focal point in the realm of computer vision, and within this domain, vehicle detection stands out as a fundamental yet challenging task due to the diverse appearance attributes of vehicles and their varying states during the detection process \cite{Lee2005}.

In this study, established models like YOLOv5, Faster-RCNN-MobileNetv3, and EfficientDet, as documented in literature, were employed for object detection. The KITTI dataset was employed to compare the performance of these models.
 
Additionally, this study introduces MelNet, a novel State-of-The-Art model, which is juxtaposed against the aforementioned models. This model, inspired by YOLOv3, was exclusively trained using the KITTI dataset.

The novel contributions of this paper are:
\begin{itemize}
    \item  A novel CNN model (MelNet) for object detection. 
    \item A utilization of batch normalization in the network architecture.
    \item A detailed validation of our deep learning model (MelNet) on the KITTI dataset.
    \item  A comparative analysis
of our model (MelNet) against other state-of-the-art models.
\end{itemize}

\section{Related Works}
\label{sec:1}
In addressing congested traffic locations, Lowe \cite{Lowe99} devised a HOG-based approach. Meanwhile, Lienhart and Maydt \cite{lienhart2002extended} utilized a hybrid DNN for the detection and classification of objects, incorporating non-negative matrix factorization (NMF) for feature extraction and compression. To tackle abrupt illumination changes and camera vibrations, Bodla et al. \cite{bodla2017soft} employed a shadow removal approach and a background Gaussian Mixture Model. This enabled the identification, tracking, and classification of vehicles into four groups, with the Kalman filter employed for tracking. Harsha and Anne \cite{Harsha2016} introduced a deep Convolutional activation feature for vehicle identification and classification, known as DeCAF. Visual features were extracted, and the accuracy of various methods, including large-scale sparse learning and deep CNN, was compared.

Ozkurt and Camci \cite{Ozkurt2009} introduced a model comprising three sub-models. The Moving Object Detector (MOD) was employed for object detection through the background estimation method. The Vehicle Identifier was used to classify vehicles into small, medium, and large categories using a neural network. Lastly, traffic density was gauged by estimating the number of recognized vehicles in consecutive frames. Experimental results conducted on the Istanbul Traffic Management Company (ISBAL) dataset were reported as promising by Ozkurt and Camci \cite{Ozkurt2009}. In another study, Suryanto et al. \cite{Suryanto2011} proposed a spatial color histogram model that encoded both color distribution and spatial information. The method employed a voting mechanism to identify new objects in the frame, followed by a back-projection method to extract the identified object from the background.

Simonyan and Zisserman \cite{Simonyan2014} employed a deep neural network for vehicle detection and categorization, aiming to extract high-level traits from lower-level ones. In experiments, the deep neural network exhibited superior performance in car classification, achieving a lower 3.34\% error rate compared to a regular neural network's 6.67\% error rate. For vehicle detection, He et al. \cite{Heetal2016} proposed an enhanced Gaussian mixture model incorporating background removal. The subsequent process involved feature extraction with AlexNet and SIFTS (scale-invariant feature transform), dimensionality reduction with PCA and LDA, and classification with SVM (support vector machine). Testing results indicated that, particularly at FC6 and FC7, the improved GMM with AlexNet DNN was more accurate in identifying and classifying cars. In a different approach, Huang et al.~\cite{Huangetal2017} developed a deep neural network utilizing the YOLO technique for detection and the AlexNet DNN for classification. The model addressed vehicle classification in dark images by employing scene modification, a late fusion approach, and a color transformation method to enhance DNN capabilities. Meanwhile, Jagannathan et al.\cite{JagannathanRFDS21} proposed an ensemble of deep learning methods for the detection and classification of moving vehicles. The model initially applied image processing techniques to enhance input image quality, followed by the use of steerable pyramid transform (SPT) and Weber local descriptor (WLD) for feature extraction. When simulated on two datasets (MIO-TCD and BIT vehicle dataset) of RGB images, the proposed model achieved impressive accuracies of $99.28\%$ and $99.13\%$.

Gao and Lee \cite{gao2015moving} presented a method for vehicle detection using Fast R-CNN. The approach involves capturing images through a camera, and the dataset is balanced with three classes. The model exhibits acceptable accuracy when images are clear and free from noise. Chan et al. \cite{Chanetal12} utilized sensor-based data for traffic analysis by deploying sensors alongside the road. Cascade filtering was employed to select data from multiple sources. A CNN-based model was then used for the classification and detection of vehicles. This method demonstrates flexibility to noise and achieves an accuracy of 98\%.  Yang et al. \cite{Yangetal21} introduced a feature-fused SSD with a tracking guided detection optimization (TDO) strategy, incorporating Non-Maximum Suppression (NMS) in the post-processing phase for vehicle detection. Biswas et al. \cite{Biswasetal2019} implemented SSD and MobileNet SSD for vehicle detection, determining that SSD exhibited superior performance compared to MobileNet. However, it was noted that MobileNet had a shorter inference time. In addressing the issue of an imbalanced dataset, Liu et al.\cite{LiuLL18} employed a semi-supervised model, integrating a Deep Neural Network (DNN) with Generative Adversarial Nets (GAN) for data augmentation.

By employing frame difference, Chen et al.~\cite{ChenLXJYF17} introduced a framework for the recognition of moving automobiles. A binary frontal view was derived through an asymmetrical filter applied to the car's front image, and the car model was identified using a three-layer limited Boltzmann machine in the context of deep learning. In a separate study, Xie et al.~\cite{XieGDTH17} focused on a vision-based system designed for the identification of preceding cars on a highway, accommodating diverse conditions such as low illumination and varying weather conditions. 

Yılmaz et al.~\cite{Yilmazetal2018} conducted a study where they employed R-CNN and Faster R-CNN deep learning techniques to train a vehicle detector. Positive classifications for the box class were assigned to region proposals with an intersection over union (IoU) overlap value of 0.5 or higher, while others were considered negative. During each stochastic gradient descent (SGD) iteration, they appropriately selected 32 positive windows and 96 background windows to create a mini batch of 128 instances. The training algorithm for the network was configured employing Stochastic Gradient Descent with Momentum (SGDM) and an initial learning rate of 0.001. The first dataset, known as the Matlab Vehicle Dataset, encompasses approximately 350 images, while the second dataset, sourced from the Stanford Vehicle Dataset, comprises around 1000 images. Specifically, the mAP values were approximately 0.73 and 0.76 for Faster R-CNN, and 0.64 and 0.65 for R-CNN, respectively \cite{Yilmazetal2018}.

Chen et al.~\cite{ChenEmbeddedSR2019} harnessed an enhanced YOLO network for real-time vehicle detection on embedded systems. The overall network architecture is characterized by its simplicity and clarity. The end-to-end nature of the network architecture contributes to faster detection speeds compared to other network model structures\cite{ChenEmbeddedSR2019}. For the trained dataset samples, they selected 8,000 images from the nuScenes autopilot dataset as a training set. Unlike conventional vehicle detection, there are four categories for detection categories. They are the front of the car, the rear of the car, the side of the car, the body of the car. By comparison, the overall performance of the modified network model is improved compared to YOLOv3-tiny. The optimized YOLOv3-live has enhanced feature extraction and learning capabilities, and has higher detection accuracy for both the “front” and the “back”.  Quantify the YOLOv3-live network model and test its mAP and FPS on embedded devices, The quantified and unquantified results are 69.79\% and 84.15\% \cite{ChenEmbeddedSR2019}. Although the accuracy of the model is reduced after being quantified, the detection accuracy can still meet the practical requirements. At the same time, the detection speed is greatly improved after being quantified, making YOLOv3-live satisfied with the application conditions of real-time vehicle condition detection\cite{ChenEmbeddedSR2019}.

Nguyen~\cite{Nguyen2019} introduced an advanced framework that extends the capabilities of Faster R-CNN for efficient vehicle detection. The first key modification involves the utilization of the MobileNet architecture to construct the fundamental convolutional layer within the Faster R-CNN framework. Furthermore, a significant enhancement is made to the post-processing step. The traditional Non-Maximum Suppression (NMS) algorithm, which typically follows the region proposal network in the original Faster R-CNN, is replaced with the soft-NMS algorithm. Soft-NMS offers a more refined approach to handling overlapping bounding boxes, leading to improved detection precision. These innovations collectively contribute to the enhanced framework's effectiveness in vehicle detection, paving the way for more rapid and accurate results. Through experimental evaluations conducted on the KITTI vehicle dataset and the LSVH dataset, the proposed approach showcases improved performance in terms of both detection accuracy and inference time when compared to the original Faster R-CNN. Specifically, the proposed method exhibits a performance enhancement of 4\% on the KITTI test set and 24.5\% on the LSVH test set, compared to the original Faster R-CNN framework\cite{Nguyen2019}. The AP values with soft-NMS experience increments of 1.33\%, 0.6\%, and 0.01\% within the "easy," "moderate," and "hard" groups, respectively, when compared to the original Faster R-CNN \cite{Nguyen2019}. The results indicate that the proposed method enhances performance compared to the Faster R-CNN framework, showing improvements of 2.49\%, 5.92\%, and 3.6\% in the "easy," "moderate," and "hard" categories, respectively. Moreover, when contrasted with the SSD framework, the proposed algorithm achieves enhancements of 11.49\%, 23.8\%, and 18.55\% in the corresponding categories. Importantly, in terms of processing efficiency, the proposed approach demonstrates rapid processing, completing an image in 0.15 seconds, while the original Faster R-CNN framework takes up to 2 seconds\cite{Nguyen2019}. 

Jamiya and Rani \cite{Jamiya2021} introduced LittleYOLO-SPP, where features were extracted through pooling layers with varying scales. To enhance network detections, Mean Square Error (MSE) and Generalized IoU (GIoU) functions were applied for bounding box regression. Jamiya and Rani \cite{Jamiya2021} asserted that the proposed model exhibited real-time vehicle detection with high accuracy. In a different approach, Zhu et al. \cite{Zhuetal21} proposed a multi-sensor multi-level enhanced CNN, considering hybrid realistic scenes involving scales, occlusion, and illumination to enhance vehicle detection capabilities. The model comprised a LiDAR Image composite module and an enhanced inference head. Experimental results indicated that the proposed model achieved reliable and accurate detection results. Conducting an extensive literature survey, AlZu’bi and Jararweh \cite{alzu2020data} focused on technologies related to autonomous vehicles, particularly those utilizing Intelligent Transportation System (ITS) techniques. The review integrated and synthesized information on monitoring autonomous vehicles through data fusion from various sources.

MelNet stands out as a cutting-edge, single-stage object detection model that leverages the power of deep learning technology. Unveiling its prowess in detecting objects across two distinct scales, MelNet represents a state-of-the-art solution in the field. Harnessing the capabilities of deep learning, this model showcases its efficiency and accuracy in identifying and localizing objects within a diverse range of scales, making it a formidable contender in modern object detection applications.

\section{Methodology}
\label{sec:3}
Various state-of-the-art models have been developed for addressing Object Detection challenges, broadly categorized into two-stage models and single-stage models.

In this research, a novel state-of-the-art model named MelNet is introduced. Additionally, the following models were trained and evaluated on the KITTI dataset: YOLOv5, EfficientDet, and Faster R-CNN-MobileNetv3.

MelNet is categorized as a single-stage model and draws inspiration from YOLOv3 \cite{RedmonetalYOLO2016}. Unlike the aforementioned models, which were trained on large datasets like ImageNet, COCO, and Pascal-VOC, MelNet was exclusively trained on the KITTI dataset. The study provides a comparative analysis of MelNet against other established state-of-the-art models.

\begin{figure}[t]
\centering
\includegraphics[width=0.94\linewidth]{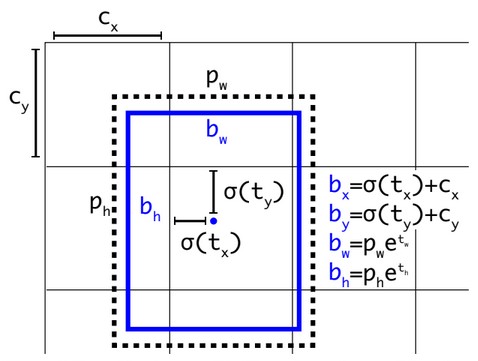}
\caption{Bounding boxes with dimension priors and location prediction \cite{Heetal2015}.}
\label{fig:1}       
\end{figure}

The network leverages dimension clusters to predict bounding boxes, a concept introduced by Redmon and Farhadi \cite{Redmon2018}. For each bounding box prediction, the network calculates four coordinates: b\textsubscript x, b\textsubscript y, b\textsubscript w, and b\textsubscript h, which represent the center coordinates, width and height dimensions (see Figure~\ref{fig:1}).

In the process of bounding box prediction, a threshold value of 0.5 is established. If the predicted bounding box doesn't achieve a satisfactory level of accuracy when compared to the corresponding ground truth bounding box, the prediction is disregarded and not taken into account. This threshold-based filtering mechanism helps ensure that only reliable bounding box predictions are considered for further evaluation and analysis.

The activation function of choice for the designed model is the Leaky ReLU. While the standard ReLU function zeros out negative inputs, Leaky ReLU introduces a small slope for negative values, thereby allowing a fraction of the negative input values to pass through. This characteristic provides a more flexible approach to handling negative inputs, ultimately contributing to the network's ability to capture diverse patterns and nuances in the data.

Furthermore, MelNet utilizes batch normalization in its architecture. Batch normalization is a prevalent normalization method applied extensively in deep learning architectures.

In MelNet, the process involves predicting boxes at two distinct scales and deriving features from these scales using a concept similar to that of feature pyramid networks. The ultimate layer of the architecture predicts a 3-dimensional tensor that encodes information about bounding boxes, objectness, and class predictions. In the context of this study, utilizing the KITTI dataset, the network forecasts 3 boxes at each scale, resulting in a tensor size of N × N × [2 × (4 + 1 + 9)], where the dimensions account for the 4 bounding box offsets, 1 objectness prediction, and 9 class predictions.

Continuing the approach of YOLOv3, the procedure involves taking the feature map from two layers back and increasing its resolution by a factor of 2 through upsampling. Additionally, a feature map obtained from an earlier stage in the network is combined with the upsampled features by employing concatenation. This design is replicated once more to forecast boxes for the ultimate scale, with the distinction that the scale is now doubled in size.

\begin{figure}[t]
\centering
\includegraphics[width=0.94\linewidth]{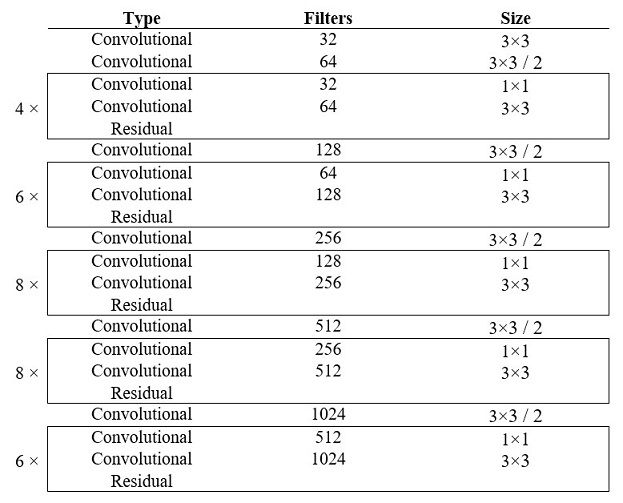}
\caption{MelNet layers.}
\label{fig:2}       
\end{figure}

\begin{figure}[t]
\centering
\includegraphics[width=0.94\linewidth]{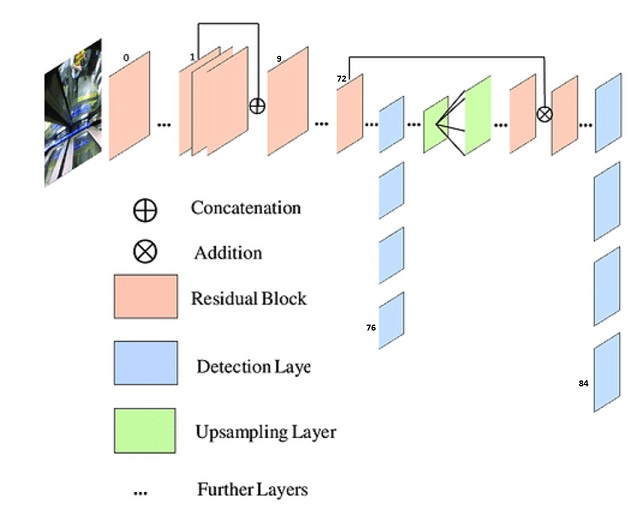}
\caption{MelNet architecture.}
\label{fig:3}       
\end{figure}

Like YOLOv3, this model also integrates a residual network. The newly introduced network utilizes a series of consecutive $3 \times 3$ and $1 \times 1$ convolutional layers, amounting to a total of 70 convolutional layers. The arrangement of layers in MelNet is outlined in Figure \ref{fig:2}, while the architecture of MelNet is visually represented in Figure \ref{fig:3}.

\section{Materials and Experiments}
\label{sec:4}
This study encompasses a variety of materials and methods, encompassing datasets as well as several Python and Object Detection libraries. The subsequent section dissects these materials and experiments, elaborating on each individually through sub-headings.
\subsection{Dataset Selection}
\label{sec:5}
\begin{figure}[t]
\centering
\includegraphics[width=0.94\linewidth]{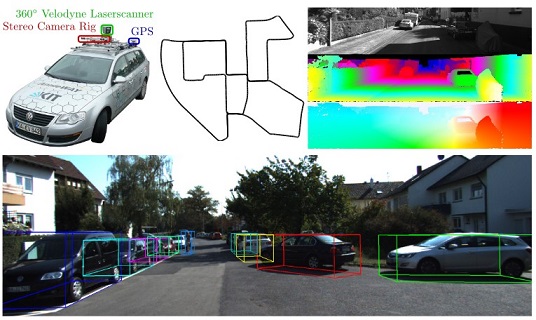}
\caption{The platform is equipped with sensors in the top-left corner. The trajectory is derived from our visual odometry benchmark and is displayed in the top-center. The top-right corner shows the disparity and optical flow map. Finally, the bottom section displays the 3D object labels \cite{GeigerAreWR2012}.}
\label{fig:kittiplatform}       
\end{figure}

The KITTI dataset, introduced by Geiger et al. \cite{GeigerAreWR2012}\cite{Geigeretal2013}. in 2012 and 2013, holds a prominent position as a widely used resource in the domains of mobile robotics and autonomous driving. This dataset encompasses various categories including Stereo, Flow, Odometry, Object (2D and 3D objects), and Tracking.
In this study, the focus is on the 2D object category within the KITTI dataset. This category comprises 7418 training images and 7518 test images, encompassing a total of 80,256 labeled objects. All images are in color and are saved in PNG format. The dataset consists of nine distinct classes, namely Car, Van, Truck, Pedestrian, Cyclist, Tram, Misc, and Don't Care (see Figure~\ref{fig:kittiplatform}). 

For the training images, they were divided into an 80\% training set and a 20\% validation set. Furthermore, the KITTI dataset employs its own bounding box format (xmin, ymin, xmax, ymax), which was converted to the YOLO bounding box format (x-center, y-center, width, height). To achieve these tasks, a straightforward Python application was developed and utilized.
\subsection{Implementation Details}
\label{sec:6}
MelNet was developed using the PyTorch library within the Python programming language. In the preprocessing phase of image data, the OpenCV, PIL, and NumPy libraries were harnessed. Additionally, the Matplotlib library facilitated the visualization of outcomes derived from the experimental investigations. The configuration of experimental platform in this study is described in Table~\ref{tab:platform}.
\begin{table}[t]
\caption{The configuration of experimental platform in this study.}
\centering
\begin{tabular}{ m{3.5cm} m{4cm} }
\hline\noalign{\smallskip}
\bf Computing Machine & \bf Configuration  \\
\noalign{\smallskip}\hline\noalign{\smallskip}
{\bf Operating System} & Ubuntu 20.04 \\
{\bf  GPU} & NVIDIA GeForce RTX 2060 Super (8GB) \\
{\bf GPU Acceleration Library} & CUDA and CUDNN \\
{\bf CPU} & Intel(R) Core i7\\
{\bf RAM} & 32GB\\ 
\noalign{\smallskip}\hline
\end{tabular}
\label{tab:platform} 
\end{table}

\subsection{Training Process}
\label{sec:7}
The Adam optimizer has gained extensive popularity within the realm of deep learning owing to its remarkable efficacy in optimizing neural networks. By amalgamating the merits of AdaGrad and RMSprop, Adam furnishes adaptive learning rates that are tailored to the specific parameters of the network. Through the estimation and adjustment of learning rates based on gradient moments, Adam dynamically adapts throughout the training process, leading to accelerated convergence and heightened performance.

Unlike alternative optimization techniques, Adam handles challenges such as hyperparameter sensitivity and the treatment of sparse gradients. Its versatility in accommodating varied network architectures and training scenarios, along with its resilience to noisy or sparse gradients, has rendered it a favored choice. Through proficient management of learning rates for individual parameters, Adam accelerates the convergence of neural networks, thereby enhancing overall performance across diverse tasks like image classification, natural language processing, and object detection. The MelNet training process incorporates the Adam optimizer.

MelNet's training solely relied on the KITTI dataset, without utilizing any pretrained models. Training occurred over 300 epochs, employing a batch size and worker count of 4. Input images were standardized to a size of $640 \times 640$ pixels. The learning rate was set to le-5 (1e-5), and weight decay was established at le-4 (1e-4).

The learning rate serves as a pivotal hyperparameter in machine learning algorithms, dictating the magnitude of parameter updates during training. It governs the extent to which model weights and biases are adjusted in response to the estimated error or loss on each optimization iteration.

Weight decay, also known as L2 regularization, is a prevalent technique employed to counter overfitting in machine learning, especially in neural networks. It functions by integrating a penalty term into the loss function, discouraging the proliferation of large weights in the model. This is achieved by appending a term proportional to the sum of squared weights (L2 norm) to the loss function. The objective is to encourage the model to favor smaller weights, as excessively large weights can contribute to overfitting. This regularization mechanism acts as a constraint, promoting simpler models characterized by diminished weight magnitudes.

In this investigation, MelNet partitions the input image into an $S \times S$ grid. Specifically, the initial scale detection layer dissects the input image into a grid of dimensions $20 \times 20$, while the subsequent scale detection layer breaks down the input image into a grid of dimensions $40 \times 40$.
Within object detection models, a pivotal metric is the mAP (mean Average Precision) value. This paper calculates the mAP values for various models, including methods based on handcrafted features include Histogram of Oriented Gradients (HOG) [20], Scale-Invariant Feature Transform (SIFT) [39], and Harr-like [51]. These approaches exhibit a limited feature representation. Despite their effectiveness with small data sizes and their independence from special hardware requirements, these methods necessitate feature extraction by experts in machine learning. The algorithm's efficiency is heavily contingent on the precision of the methods used for identifying and extracting features.YOLOv5, MobileNetv3, EfficientDet, and MelNet. These models were trained on the KITTI dataset to assess their performance in object detection tasks.
\subsection{Data Augmentation}
\label{sec:8}
Image data underwent augmentation through a series of operations and transformations utilizing the Albumentations library. This facilitated the application of diverse image processing techniques, including random rotations, flips, crops, and color adjustments, thereby enhancing dataset diversity.

\section{Results}
\label{sec:9}
\subsection{Metrics Evaluation Results}
\label{sec:10}
MelNet was trained on the KITTI dataset, refraining from using any preexisting backbone. Due to this approach, distinct training epochs were chosen. The designated epoch numbers for training were 50, 100, 150, 200, 250, and 300 epochs.  The training outcomes are tabulated in Table \ref{tab:2}.

\begin{figure} [t]
\centering
\includegraphics[width=1\linewidth]{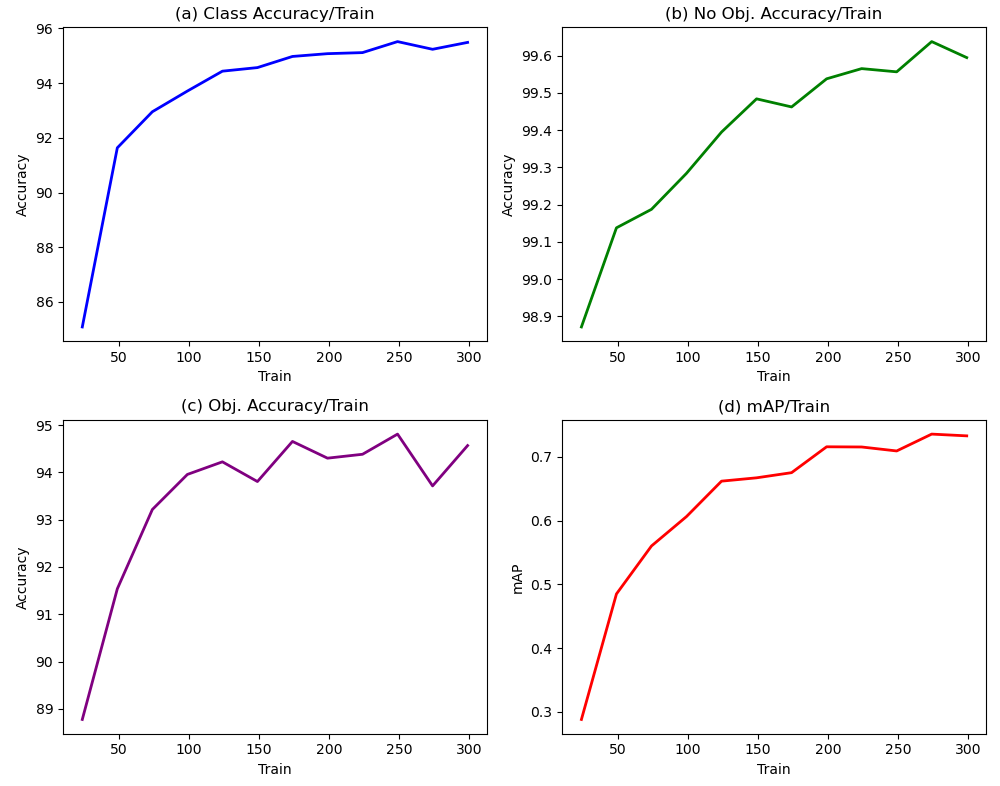}
\caption{(a) Class Accuracy result of training MelNet. (b) No Object Accuracy result of training MelNet. (c) Object Accuracy result of training MelNet. (d) mAP result of training MelNet.}
\label{fig:5}       
\end{figure}

\begin{figure} [t]
\centering
\includegraphics[width=1\linewidth]{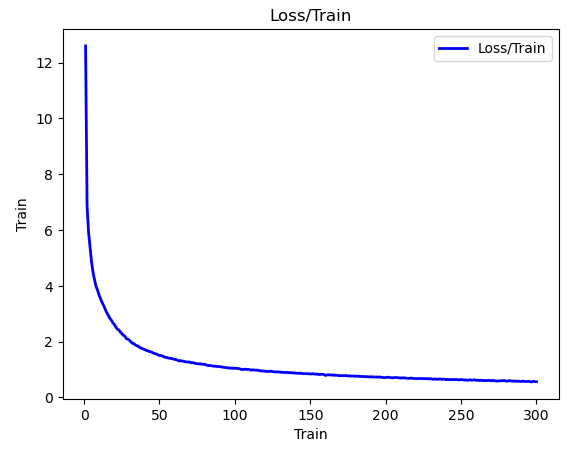}
\caption{Loss value result of training MelNet.}
\label{fig:6}       
\end{figure}

\begin{table*}[t]
\caption{MelNet results for various number of epochs.}
\begin{tabular}{ m{2cm} m{1.7cm} m{1.7cm} m{1.7cm} m{1.7cm} m{1.7cm} m{1.7cm} }
\hline\noalign{\smallskip}
\bf Metric & \bf 50 EPOCH & \bf 100 EPOCH & \bf 150 EPOCH & \bf 200 EPOCH & \bf 250 EPOCH & \bf 300 EPOCH  \\
\noalign{\smallskip}\hline\noalign{\smallskip}
{\bf mAP} & 0.49 & 0.61 & 0.69 & 0.72 & 0.72 & 0.73 \\
{\bf LOSS} & 1.51 & 1.06 & 0.84 & 0.71 &0.63 & 0.57 \\
{\bf Class Acc.} & 91.36 & 93.94 & 94.91 & 95.23 & 95.25 & 95.52 \\
{\bf Obj. Acc.} & 91.90 & 93.70 & 94.12 & 95.28 & 94.48 & 94.74 \\
{\bf No Obj. Acc.} & 99.08 & 99.33 & 99.47 & 99.49 & 99.56 & 99.59 \\
\noalign{\smallskip}\hline
\end{tabular}
\label{tab:2} 
\end{table*}
As indicated in Table \ref{tab:3}, the highest mAP value of 0.770 is attributed to Faster RCNN-MobileNetv3. Furthermore, noteworthy mAP values of 0.749, 0.732, and 0.667 are achieved by YOLOv5, MelNet, and EfficientDet, respectively.
\begin{table}[t]
\caption{A comparison of mAP results on KITTI Dataset.}
\begin{tabular}{ m{2cm} m{2cm} m{1.2cm} m{1.2cm}}
\hline\noalign{\smallskip}
\bf Model & \bf Input Size & \bf Dataset & \bf mAP \\
\noalign{\smallskip}\hline\noalign{\smallskip}
 MobileNetv3 & 640 × 640 & KITTI & {\bf0.770} \\
 YOLOv5  & 640 × 640 & KITTI & 0.749  \\
{\bf MelNet} & 640 × 640 & KITTI  & {\bf 0.732} \\
 EfficientDet & 640 × 640 & KITTI & 0.667  \\
\noalign{\smallskip}\hline
\end{tabular}
\label{tab:3} 
\end{table}
In Figures \ref{fig:5} and \ref{fig:6}, the outcomes pertaining to mAP value, Loss value, Class Accuracy, No Object Accuracy, and Object Accuracy resulting from training MelNet over 300 epochs are depicted.
MelNet's main architecture encompasses a total of 70 layers, which play a crucial role in yielding improved accuracy outcomes. As it's shown in Figure \ref{fig:5}, the accuracy results are depicted as remarkably satisfactory. 

As it's seen in Figure \ref{fig:6}, the visualization demonstrates the success of the training process, with no signs of overfitting. The Loss value consistently decreases with each epoch, culminating in a final value of 0.574.

\begin{figure} [t]
\centering
\includegraphics[width=1\linewidth]{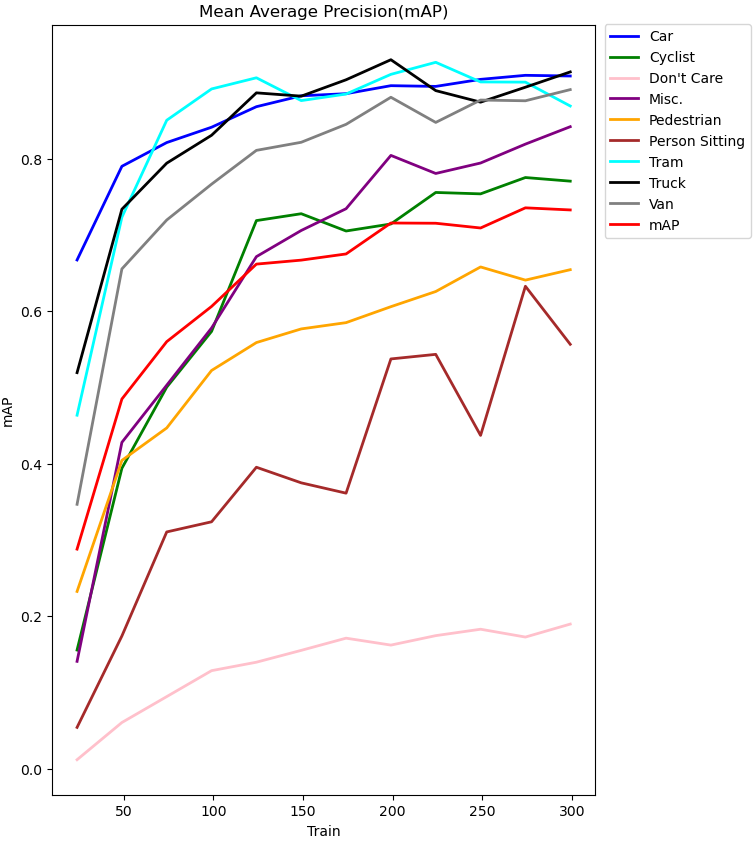}
\caption{mAP results of our MelNet.}
\label{fig:7}       
\end{figure}

The mAP value of each class is shown in Figure \ref{fig:7}. As you can see the lowest mAP value belongs to "Don't Care" class and the highest value belongs to "Truck" class.

\subsection{Prediction Result}
\label{sec:11}

\begin{figure} [t]
\centering
\includegraphics[width=1\linewidth]{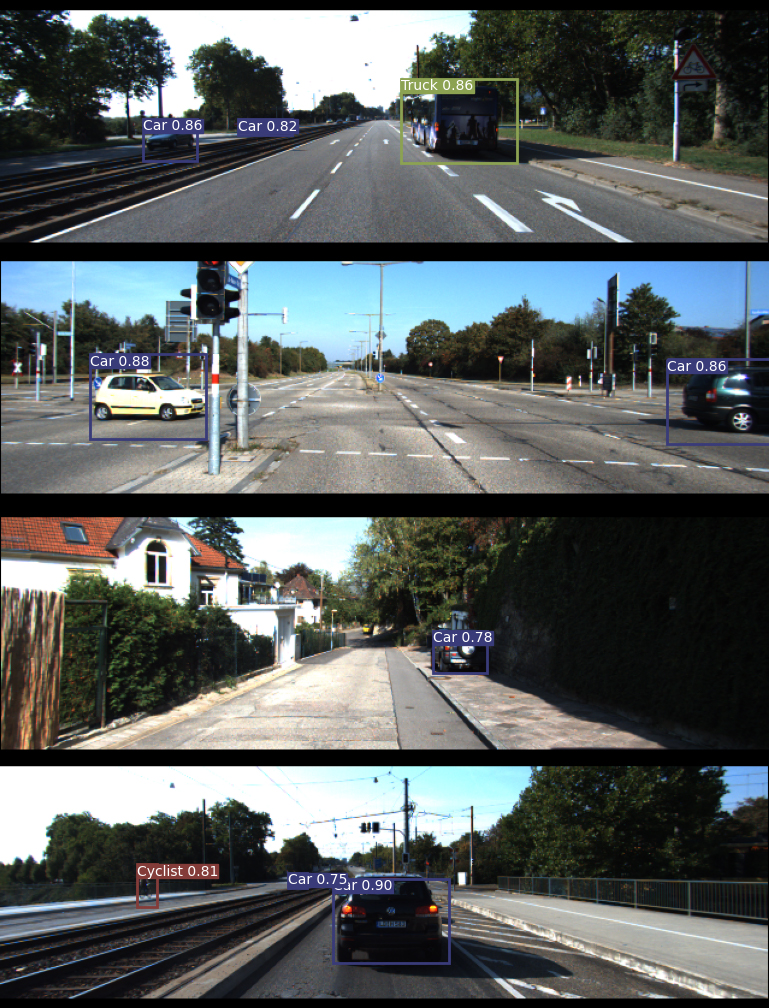}
\caption{Prediction results of our MelNet.}
\label{fig:8}       
\end{figure}

\begin{figure} [t]
\centering
\includegraphics[width=1\linewidth]{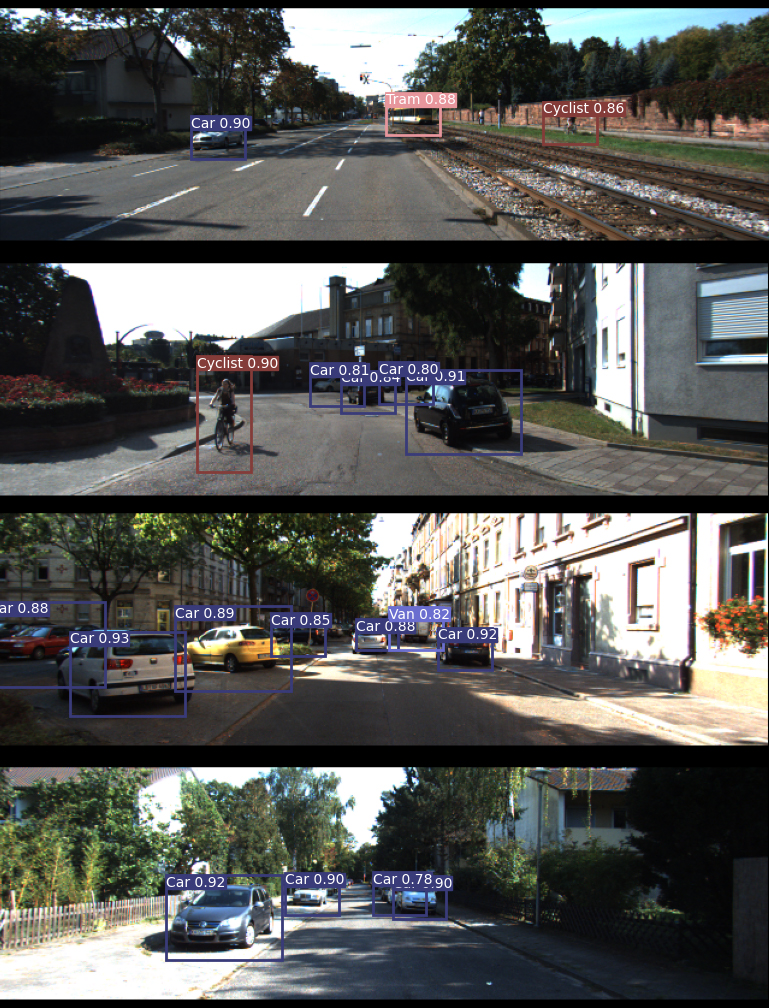}
\caption{Prediction results of our MelNet.}
\label{fig:9}       
\end{figure}

Figure \ref{fig:8} and \ref{fig:9} show the prediction results of MelNet. Investigating the performance of MelNet, it is evident from Figure \ref{fig:8} and \ref{fig:9} that the model excels in object detection. However, there is room for improvement in refining the precision of the bounding boxes. While MelNet demonstrates an impressive ability to accurately identify objects, optimizing the bounding box localization would further enhance its overall detection capabilities.

\subsection{Training Time Comparison Results}
\label{sec:12}
Table \ref{tab:4} shows the training time comparison over 50 epochs between these models.
\begin{table*}[t]
\caption{A training time comparison of various deep learning algorithms.}
\begin{tabular}{ m{2.2cm} m{2.2cm} m{2.2cm} m{2.2cm} m{2.2cm} m{2.3cm} }
\hline\noalign{\smallskip}
\bf Model & \bf Epoch Number & \bf Input Size & \bf Dataset & \bf Batch Size & \bf Training Time (h)  \\
\noalign{\smallskip}\hline\noalign{\smallskip}
{\bf MobileNetv3} & 50 & 640 × 640 & KITTI & 4 & 7.983 \\
{\bf MelNet} & 50 & 640 × 640 & KITTI & 4 & 8.7 \\
{\bf YOLOv5} & 50 & 640 × 640 & KITTI & 4 & 9.32 \\
{\bf EfficientDet} & 50 & 640 × 640 & KITTI  & 4 &11.666 \\
\noalign{\smallskip}\hline
\end{tabular}
\label{tab:4} 
\end{table*}
As indicated in Table \ref{tab:4}, the model with the quickest training time is belongs to MobileNetv3, clocking in at 7.983 hours. On the other end of the spectrum, EfficientDet took the longest training time, spanning 11.666 hours. MelNet is in the second place, requiring approximately 8.7 hours for training. It's noteworthy that all these models were trained on the same computer. The training time variations for each model were influenced by the batch size values, which are detailed in Table \ref{tab:4}.
\subsection{Number of Layers Comparison}
\label{sec:13}
Table \ref{tab:5} shows number of layers of various deep learning algorithms used in this study. As you can see in  Table \ref{tab:4} and Table \ref{tab:5}, MelNet has a good performance with lowest number of layers. This also shows that MelNet can be implemented in real-time object detection systems.
\begin{table}[t]
\centering
\caption{A comparison of number of layers of various deep learning algorithms.}
\begin{tabular}{ m{3cm} m{3cm} }
\hline\noalign{\smallskip}
\bf Model & \bf Number of Layers \\
\noalign{\smallskip}\hline\noalign{\smallskip}
{\bf MelNet} & 72 \\
{\bf YOLOv5} & around 130  \\
{\bf MobileNetv3} & around 130 \\
{\bf EfficientDet} & around 153 \\
\noalign{\smallskip}\hline
\end{tabular}
\label{tab:5} 
\end{table}
\section{Conclusion and Future Works}
\label{sec:14}
Numerous models and techniques have emerged in the realm of object detection, encompassing both single-object detectors and two-stage object detectors. Within this study, YOLOv5, EfficientDet, Faster-RCNN-MobileNetv3, and the novel proposed model - MelNet - were trained on the KITTI dataset and subjected to a comprehensive comparison, with mAP value serving as the central evaluation metric.

The findings underline the effectiveness of transfer learning in certain scenarios. Pretrained models, trained on prominent datasets like ImageNet, COCO, and Pascal VOC, tend to exhibit superior performance due to their broader training data.

Additionally, the results highlight that designing a tailored model for a specific context and training it on a focused dataset can also yield satisfactory results. This study underscores the capacity of MelNet, trained solely on the KITTI dataset, to outperform EfficientDet. Upon training, MelNet's results are in close proximity to those of other pretrained models.

In future endeavors, the primary focus could involve training MelNet on larger datasets such as ImageNet or COCO. This approach would leverage the benefits of transfer learning, enabling the pretrained model to be fine-tuned on the KITTI dataset for further performance enhancement. Additionally, there is potential for upgrading and optimizing MelNet by incorporating state-of-the-art methods and techniques~\cite{Gok2023SIU}. Furthermore, we are also interested in implementing our novel deep learning algorithm for representing Bidirectional Reflectance Distribution Functions (BRDFs)~\cite{Ozturk2006EGUK,Kurt2007MScThesis,Ozturk2008CG,Kurt2008SIGGRAPHCG,Kurt2009SIGGRAPHCG,Kurt2010SIGGRAPHCG,Ozturk2010GraphiCon,Ozturk2010CGF,Bigili2011CGF,Bilgili2012SCCG,Tongbuasirilai2017ICCVW,Kurt2019DEU,Akleman2024arXiv}, Bidirectional Scattering Distribution Functions (BSDFs)~\cite{WKB12,Ward2014MAM,Kurt2014WLRS,Kurt2016SIGGRAPH,Kurt2017MAM,Kurt2018DEU}, Bidirectional Surface Scattering Reflectance Distribution Functions (BSSRDFs)~\cite{Kurt2013TPCG,Kurt2013EGSR,Kurt2014PhDThesis,Onel2019PL,Kurt2020MAM,Kurt2021TVC,Yildirim2024arXiv} and multi-layered materials~\cite{WKB12,Kurt2016SIGGRAPH,Mir2022DEU} in computer graphics.


%
%



\bibliographystyle{spmpsci}      
\bibliography{arXiv24_MelNet_References}   

\end{document}